\documentclass[10pt,twocolumn,letterpaper]{article}

\usepackage{iccv}
\usepackage{times}
\usepackage{epsfig}
\usepackage{graphicx}
\usepackage{amsmath}
\usepackage{amssymb}
\usepackage{algorithm}
\usepackage{algorithmic}
\usepackage{booktabs}
\usepackage{array}
\newcolumntype{C}[1]{>{\centering\arraybackslash}m{#1}}
\newcommand{\stress}[1]{\textbf{{#1}}}
\def\etal{\emph{et al}\onedot}

\usepackage[breaklinks=true,bookmarks=false]{hyperref}

\iccvfinalcopy 

\def\iccvPaperID{****} 

\begin{document}

\title{MASC: Multi-scale Affinity with Sparse Convolution \\ for 3D Instance Segmentation\\
\large{Technical Report}}

\author{Chen Liu\\
Washington University in St. Louis\\
{\tt\small chenliu@wustl.edu}
\and
Yasutaka Furukawa\\
Simon Fraser University\\
{\tt\small furukawa@sfu.ca}
}

\maketitle

\begin{abstract}
We propose a new approach for 3D instance segmentation based on sparse convolution and point affinity prediction, which indicates the likelihood of two points belonging to the same instance. The proposed network, built upon submanifold sparse convolution~\cite{graham20183d}, processes a voxelized point cloud and predicts semantic scores for each occupied voxel as well as the affinity between neighboring voxels at different scales. A simple yet effective clustering algorithm segments points into instances based on the predicted affinity and the mesh topology. The semantic for each instance is determined by the semantic prediction. Experiments show that our method outperforms the state-of-the-art instance segmentation methods by a large margin on the widely used ScanNet benchmark~\cite{dai2017scannet}. We share our code publicly at~\href{https://github.com/art-programmer/MASC}{https://github.com/art-programmer/MASC}.
\end{abstract}

\section{Introduction}

\begin{figure*}[t]
	\centering
    \includegraphics[width=\linewidth]{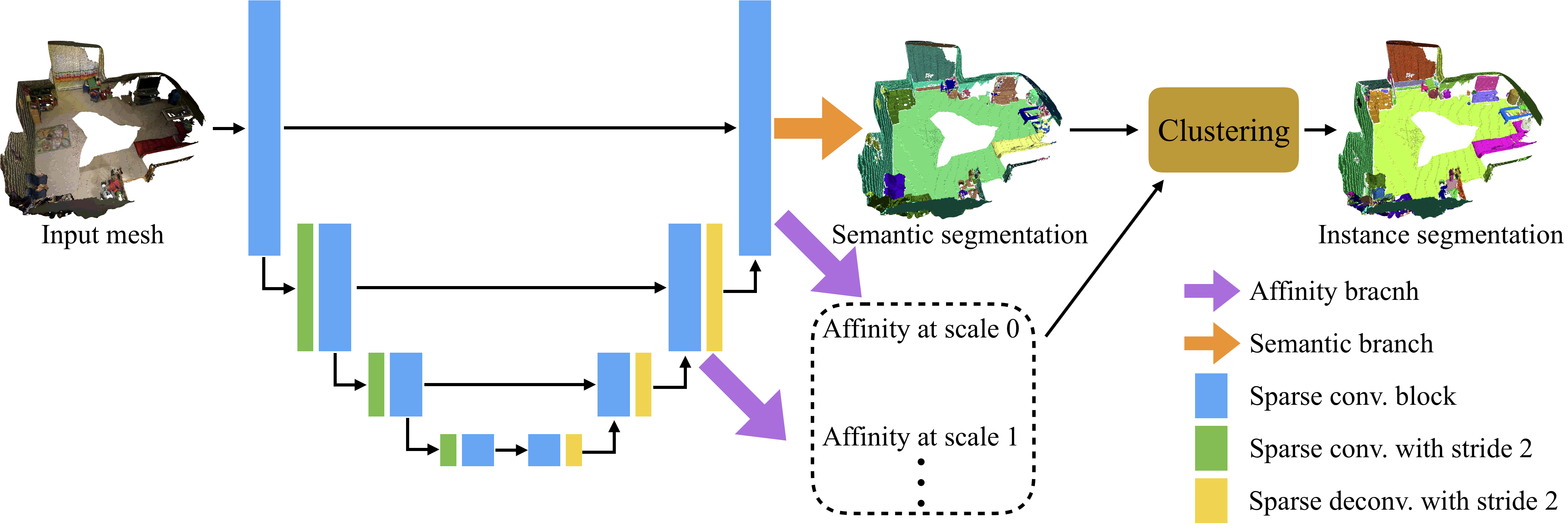}
\caption{We use a U-Net architecture with submanifold sparse convolutions~\cite{graham20183d} to process the entire point cloud of an indoor scene and predict semantic scores for each point as well as the affinity between neighboring voxels at different scales. A simple yet effective clustering algorithm groups points into instances based on the predicted affinity and the mesh topology.}
	\label{fig:pipeline}
\end{figure*}

The vision community has witnessed tremendous progress in 3D data capturing and processing techniques in recent years. Consumer-grade depth sensors enable researchers to collect large-scale datasets of 3D scenes~\cite{dai2017scannet,armeni2017joint,hua2016scenenn}. The emergence of such datasets empowers learning-based methods to tackle a variety of 3D tasks, such as object recognition, part segmentation and semantic segmentation. Among them, 3D instance segmentation is an important yet challenging task as it requires accurate segmentation of 3D points based on instances without a fixed label set. In this paper, we propose a simple yet effective method to learn the affinity between points based on which points are clustered into instances.

The irregularity of 3D data poses a challenge for 3D learning techniques. Volumetric CNNs~\cite{wu20153d,qi2016volumetric,maturana2015voxnet} are first explored as they are straightforward extensions of their successful 2D counterparts. Various techniques are proposed to reduce the cost of expensive 3D convolutions. OctNet~\cite{riegler2017octnet} and O-CNN~\cite{wang2017cnn} utilize the sparsity of 3D data via the octree representations to save computation. Other methods~\cite{klokov2017escape,qi2017pointnet} process 3D point cloud directly without voxelization. While showing promising results, these methods lack the ability of modeling local structures of the input point cloud. Later methods~\cite{qi2017pointnet++,huang2018recurrent,liu2018floornet} model local dependencies with various tricks, but the number of points processed by the network is still quite limited (e.g., $4,096$). A point cloud of an indoor space usually contains much more number of points, which means the network can only process a slice of the input point cloud at each time, which disables global reasoning of the space. Recently, Graham~\etal~\cite{graham20183d} propose an super-efficient volumetric CNN based on submanifold sparse convolution~\cite{graham2017submanifold} to process the entire point cloud of an indoor scene, which achieves promising results on the semantic segmentation task. In this paper, we adopt sparse convolution and propose a clustering algorithm based on learned multi-scale affinities to tackle the 3D instance segmentation problem.

Instance segmentation is more challenging than semantic segmentation as it requires the additional reasoning of objects. Instance segmentation methods can be categorized into two groups, proposal-based approaches and proposal-free approaches. Proposal-based approaches builds system upon object detection and append segmentation modules after bounding box proposals. Inspired by the recent success of Mask R-CNN~\cite{he2017mask} on 2D instance segmentation, 3D-SIS~\cite{hou20183d} develops a proposal-based system which achieves the state-of-the-art performance on 3D instance segmentation evaluated using the ScanNet benchmark~\cite{dai2017scannet}. GSPN~\cite{yi2018gspn} presents a generative model for generating proposals. FrustumNet~\cite{qi2018frustum} un-project 2D proposals to 3D space for the segmentation network to process. On the other hand, proposal-free methods cluster points into instances based on the similarity metrics. SGPN~\cite{wang2018sgpn} trains a network to predict semantic labels, a similarity matrix between all pairs of points, and point-wise confidence for being a seed point, from which instance segmentation results are generated. However, the similarity between most pairs of points is not informative and introduces unnecessary learning challenge, limiting the network to process only $4,096$ points at each time. We address this issue by predicting only similarity between neighboring points at multiple scales, as did in~\cite{liu2018affinity} for 2D instance segmentation, and develops a clustering algorithm to group points based on the learned local similarity and the input mesh topology.

\section{Methods}
As shown in Fig.~\ref{fig:pipeline}, we use the same U-Net architecture~\cite{ronneberger2015u} with submanifold sparse convolution used in~\cite{graham20183d} for semantic segmentation. The sparse U-Net first voxelized the entire point cloud from each ScanNet scene, with each point represented by its coordinate, color, and local surface normal. Following~\cite{graham20183d}, we set the voxel size as $2cm \times 2cm \times 2cm$ and use $4,096 \times 4,096 \times 4,096$ voxel grids so that the entire scene can be voxelized sparsely and each voxel usually contains at most 1 point. After voxelization, submanifold sparse convolution layers with skip connections generate feature maps of different spatial resolutions. We append a semantic branch with one fully connected layer to predict point-wise semantics and add multiple affinity branches at different scales to predict the similarity scores between an active voxel and its 6 neighbors. We denote the finest scale, which has $4,096 \times 4,096 \times 4,096$ voxels, as scale 0 and scale $s$ has resolution $\frac{4,096}{s^2} \times \frac{4,096}{s^2} \times \frac{4,096}{s^2}$.

\begin{algorithm}[h]
\caption{The clustering algorithm based on multi-scale affinity}
\begin{algorithmic} 
\REQUIRE Voxel affinity $A^s(p, q)$ at each scale, input mesh $(V, E)$
\RETURN Clustering result for $V$
\REPEAT
\STATE Initialize $A(V_i, V_j)$\\
\IF{$(V_i, V_j) \in E$}
\STATE $A(V_i, V_j) \leftarrow \mathrm{avg}_{s}(\mathrm{avg}_{p \in V_i, q \in V_j}(A^s(p, q)))$
\ELSE
\STATE $A(V_i, V_j) \leftarrow 0$
\ENDIF
\STATE Map $V_i$ to a neighbor $M(V_i)$ with high similarity \\
\IF{$\mathrm{max}_{V_j}(A(V_i, V_j)) > 0.5$}
\STATE $M(V_i) \leftarrow \mathrm{argmax}_{V_j}(A(V_i, V_j))$
\ELSE
\STATE $M(V_i) \leftarrow V_i$
\ENDIF
\STATE Assign $V_i$ to cluster $C_k$ if $M(V_i) \neq V_i, M(V_i) \in C_k$
\STATE Update $E \leftarrow \{(C_k, C_l) | \exists V_i, V_j : V_i \in C_k, V_j \in C_l, (V_i, V_j) \in E\}$
\STATE Update $V \leftarrow C$
\UNTIL{The update does not change anything}
\end{algorithmic}
\label{alg:clustering}
\end{algorithm}

We treat the input mesh as the initial graph $(V, E)$, and propose a clustering algorithm to group points into instances based on the multi-scale affinity field and the semantic prediction. Note that the clustering is conducted on the mesh graph while the network predicts the affinity between neighboring voxels with fixed neighborhood. So we first compute the affinity between two notes by taking the average affinity between neighboring voxel pairs connecting two nodes. At the beginning, each node contains only one point in the original point cloud which occupies one voxel and the affinity between two nodes is simply the affinity between corresponding voxels. As the node grows, a node could occupy multiple voxels in the finest scale, and if no less than $4^s$ points of the node fall inside the voxel at a higher scale $s$, we say that the node occupies this voxel. With node affinities, we map each node $V_i$ to its most similar neighbor $M(V_i)$ if the similarity between them is higher than $0.5$, and to itself otherwise. With the mapping, we can cluster nodes into groups such that every node in a group is mapped to another node in the same group and none of the node in a group is mapped to other groups. We treat each cluster as a new node and update edges correspondingly. We repeat the process until no change happens (i.e., every node is mapped to itself). The above procedure is summarized in Alg.~\ref{alg:clustering}.

Compared against the clustering algorithm in~\cite{liu2018affinity} for 2D instance segmentation, our clustering algorithm is more aggressive as it merges nodes in parallel, and has the potential of being implemented using GPU operations. After the clustering process, each instance takes the semantic label with the maximum votes from its points.

\section{Results}
\subsection{Implementation details}
We implement the proposed method using PyTorch based on the code of~\cite{graham20183d}\footnote{https://github.com/facebookresearch/SparseConvNet}. Besides voxelizing and augmenting ScanNet scenes in the same way of~\cite{graham20183d}, we add point normals to the input and use instance labels to generate affinity supervision between neighboring voxels. Two neighboring voxels at scale $s$ have a similarity score 1 if 1) each of them contains at least $4^s$ points and 2) their instance distributions among points are the same. To better predict local affinity at regions where the input mesh is sparse, we further augment the input mesh by randomly sampling 5 points inside each triangle which spans at least 2 voxel at either dimension. An edge is added between two sampled points if they are voxelized into neighboring voxels at the finest scale, and a sampled point is also connected with its closest vertex of the original triangle. This random densification also enriches the training data. We use 2 scales for the affinity prediction. We train the network on a Titan 1080ti GPU for 20 epochs.

\subsection{Quantitative evaluation}

\begin{table*}[t]
\scriptsize
\caption{Instance segmentation evaluation on the ScanNet benchmark (AP with IOU threshold $0.5$)}
\label{tbl:comparison}
  \centering
  \begin{tabular}{l|C{0.45cm}|*{20}{C{0.45cm}}}
    \toprule
    Method & avg & bath. & bed & book. & cabi. & chair & coun. & curt. & desk & door & other & pict. & refr. & show. & sink & sofa & table & toil. & wind.\\
    \midrule[1pt]
3D-SIS~\cite{hou20183d} & 0.382 & \stress{1.000} & 0.432 & 0.245 & 0.190 & 0.577 & 0.013 & 0.263 & 0.033 & 0.320 & 0.240 & 0.075 & 0.422 & \stress{0.857} & 0.117 & \stress{0.699} & 0.271 & 0.883 & 0.235 \\
\midrule
GSPN~\cite{yi2018gspn} & 0.306 & 0.500 & 0.405 & 0.311 & 0.348 & 0.589 & \stress{0.054} & 0.068 & 0.126 & 0.283 & 0.290 & 0.028 & 0.219 & 0.214 & 0.331 & 0.396 & 0.275 & 0.821 & 0.245 \\
 \midrule
SGPN~\cite{wang2018sgpn} & 0.143 & 0.208 & 0.390 & 0.169 & 0.065 & 0.275 & 0.029 & 0.069 & 0.000 & 0.087 & 0.043 & 0.014 & 0.027 & 0.000 & 0.112 & 0.351 & 0.168 & 0.438 & 0.138 \\
\midrule
\midrule
Ours & \stress{0.447} & 0.528 & \stress{0.555} & \stress{0.381} & \stress{0.382} & \stress{0.633} & 0.002 & \stress{0.509} & \stress{0.260} & \stress{0.361} & \stress{0.432} & \stress{0.327} & \stress{0.451} & 0.571 & \stress{0.367} & 0.639 & \stress{0.386} & \stress{0.980} & \stress{0.276} \\
      \bottomrule
  \end{tabular}
\vspace{-5pt}
\end{table*}

We evaluate the instance segmentation performance on the ScanNet benchmark~\cite{dai2017scannet}. To measure the confidence of each instance for the evaluation purpose, we collect all predicted instances in the training set and train a smaller network to predict if an instance is valid for the predicted label. The network encodes the point cloud of the instance using sparse convolutions and encodes the predicted label with a fully connected layer. We concatenate two features and add two fully connected layers to predict the final confidence score. In addition, we find that the clustering algorithm sometimes struggles to segment objects which are co-planar with other larger objects. So we add additional instances for semantic classes which are often planar, namely pictures, curtains, shower curtains, sinks, and bathtubs (usually contain only the bottom surfaces), by finding connected components of such classes based on the predicted semantics. Our method out-performs the state-of-the-art methods by a large margin. Table~\ref{tbl:comparison} shows the comparison against published results (with citable publications).

\subsection{Qualitative evaluation}
\begin{figure}[h]
	\centering
    \includegraphics[width=\linewidth]{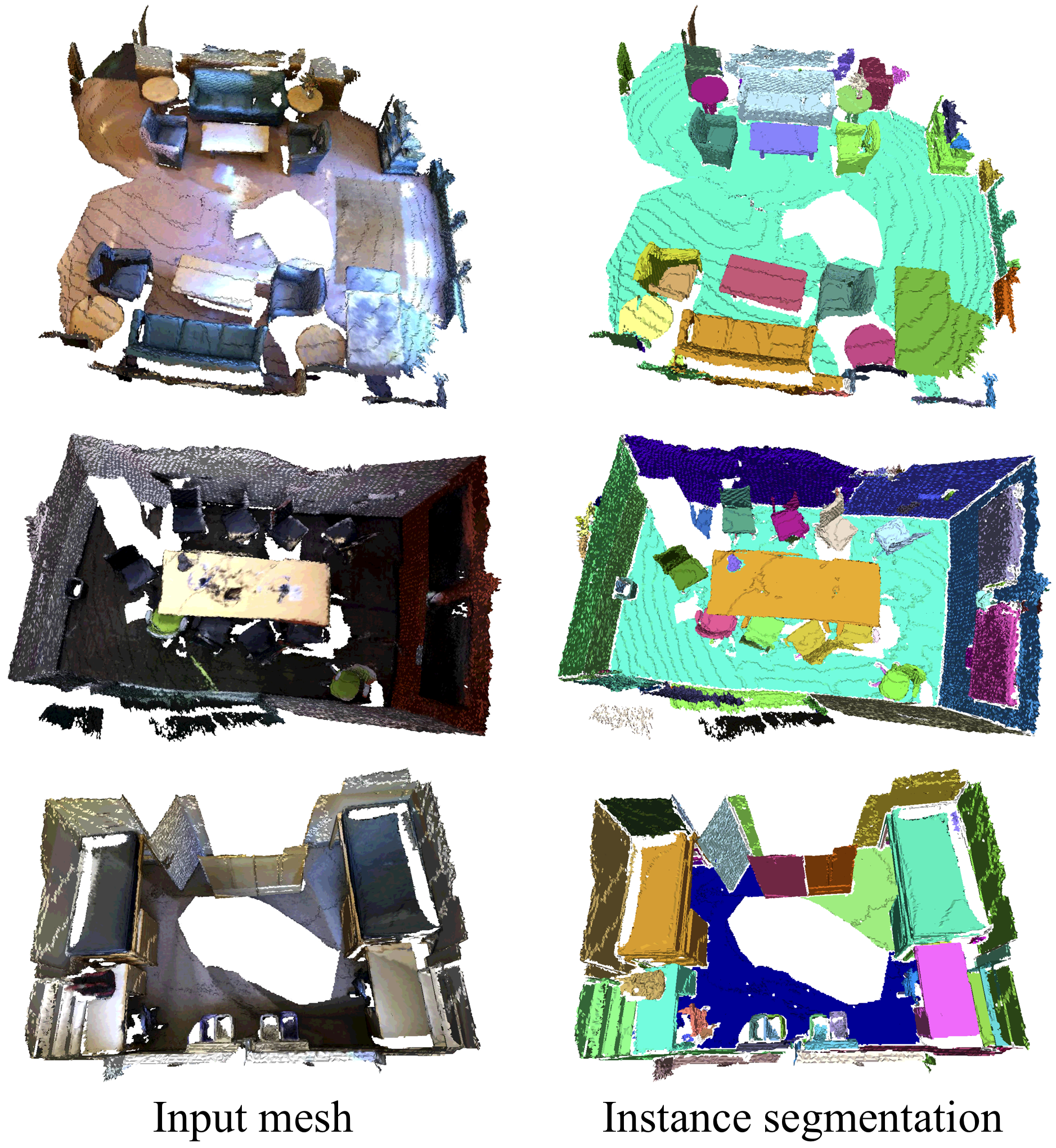}
\caption{Qualitative results for ScanNet scenes. In each row, we show the input mesh on the left and our instance segmentation result on the right.}
	\label{fig:results}
\end{figure}

We show qualitative results for ScanNet scenes in the validation set in Fig.~\ref{fig:results}.

\section{Discussion}
Though this simple method already achieves promising results, several improvements are required to further boost its performance. First, the clustering algorithm is currently implemented sequentially and thus slow. The algorithm is parallel in theory and it is possible to implement it on GPU and extend it to enable back-propagation for end-to-end training. Besides the speed issue, the effect of multi-scale is under-explored. In practice, we find using 2 scales is fast to train and achieves good performance but it is unclear the role played by each scale. With more exploration on the multi-scale affinity, it is possible to design a better clustering algorithm which uses affinity of more scales to achieve better performance. Finally, the current method sometimes fails to distinguish co-planar objects.

\section{Acknowledgement}
This research is partially supported by National Science Foundation under grant IIS 1618685, 
Natural Sciences and Engineering Research Council under grants RGPAS-2018-528049, RGPIN-2018-04197, and DGDND-2018-00006, and Google Faculty Research Award. We thank
Nvidia for a generous GPU donation.

{\small
\bibliographystyle{ieee}
\bibliography{egbib}
}

\end{document}